\documentclass[runningheads]{llncs}
\usepackage{graphicx}

\usepackage{tikz}
\usepackage{comment}
\usepackage{amsmath,amssymb} 
\usepackage{color}

\usepackage{ctable}
\usepackage{algorithm}
\usepackage{enumitem}
\usepackage{xcolor}
\usepackage{hyperref}
\usepackage[normalem]{ulem}
\usepackage{lmodern}

\usepackage{array}
\newcolumntype{P}[1]{>{\centering\arraybackslash}p{#1}}
\newcolumntype{M}[1]{>{\centering\arraybackslash}m{#1}}
\newcommand{\PreserveBackslash}[1]{\let\temp=\\#1\let\\=\temp}
\newcolumntype{C}[1]{>{\PreserveBackslash\centering}p{#1}}
\newcolumntype{R}[1]{>{\PreserveBackslash\raggedleft}p{#1}}
\newcolumntype{L}[1]{>{\PreserveBackslash\raggedright}p{#1}}
\usepackage{float}
\usepackage{subcaption}
\usepackage{hhline}

\def\eg{\emph{e.g.},} 
\def\ie{\emph{i.e.},}

\usepackage[accsupp]{axessibility}  

\begin{document}

\pagestyle{headings}
\mainmatter
\def\ECCVSubNumber{6138} 

\title{Revisiting Batch Norm Initialization}

\titlerunning{Revisiting Batch Norm Initialization}

\author{Jim Davis \and Logan Frank}
\authorrunning{J. Davis and L. Frank}
%
\institute{Department of Computer Science and Engineering\\
Ohio State University \\
\email{\{davis.1719, frank.580\}@osu.edu}}

\maketitle

\begin{abstract}
Batch normalization (BN) is comprised of a normalization component followed by an affine transformation and has become essential for training deep neural networks. Standard initialization of each BN in a network sets the affine transformation scale and shift to 1 and 0, respectively. However, after training we have observed that these parameters do not alter much from their initialization. Furthermore, we have noticed that the normalization process can still yield overly large values, which is undesirable for training. We revisit the BN formulation and present a new initialization method and update approach for BN to address the aforementioned issues. Experiments are designed to emphasize and demonstrate the positive influence of proper BN scale initialization on performance, and use rigorous statistical significance tests for evaluation. The approach can be used with existing implementations at no additional computational cost. Source code is available at \texttt{\href{https://github.com/osu-cvl/revisiting-bn-init}{ https://github.com/osu-cvl/revisiting-bn-init}}.
\end{abstract}

\section{Introduction}

Batch normalization (BN) \cite{ioffe2015batch} is a standard component used in deep learning, particularly for convolutional neural networks (CNNs) \cite{simonyan2014very,he2016deep,sandler2018mobilenetv2,xie2017aggregated,ding2021repvgg} where BN layers typically fall into a convolutional-BN-ReLU sequence \cite{he2016deep,huang2017densely}. BN constrains the intermediate per-channel features in a network by utilizing the statistics among examples in the same batch to normalize (or ``whiten") the data, followed by a learnable affine transformation to add further flexibility in training. It has been shown that the aggregation of information in a batch is advantageous as it eases the optimization landscape by creating smoother gradients \cite{santurkar2018does} and enables larger learning rates for faster convergence \cite{ioffe2015batch,bjorck2018understanding}. Additionally, the stochasticity from batch statistics can benefit generalization \cite{ioffe2015batch,luo2018towards}. 

Typically, the default implementation of BN initializes the affine transformation scale ($\gamma$) and shift ($\beta$) parameters to 1 and 0, respectively, with the expectation that they can adapt to optimal values during training and can undo the preceding normalization if desired \cite{ioffe2015batch}. It is common to take this default initialization ``as is", with extensions of BN typically adding additional parameters and computations \cite{jia2019instance,li2020attentive,liang2020instance,gao2021representative} that increase the model complexity.

By inspection of multiple trained BNs, we have observed that the affine parameters often do not change much from their initial values. So either the affine transformation is hardly needed (staying close to identity) or perhaps there is a limitation in the learning process keeping these parameters from  reaching more optimal values. It is unlikely the affine parameters are optimal near identity settings, therefore we question whether the default affine transformation initialization and update approach are appropriate. 

In this work, we present a thorough analysis of BN to address the aforementioned hypothesis and provide a new approach for the initialization and updating of BN to enhance overall performance. We will show that 1) reducing the initialization value of the BN scale parameter and 2) decreasing the learning rate on the BN scale parameter can together lead to significant improvements. These alterations to BN are straightforward adaptions to standard implementations with existing libraries (\eg\ PyTorch \cite{paszke2019pytorch}, TensorFlow \cite{tensorflow2015}) and have no impact on model complexity or training time. We additionally present a means to further apply BN to the task of input data normalization. 

Experiments are provided to compare our proposed BN method to standard BN across multiple popular benchmark datasets and network architectures, and further compare with alternative and existing methods. Notably, we conduct \textit{multiple} training runs, each using a different random seed, for every experiment to properly show statistically significant differences. Results indicate that the proposed technique can yield significant gains with no additional computational costs. Our contributions are summarized as follows:
\begin{enumerate}[noitemsep,nolistsep]
    \item A new BN initialization and update method for improving performance with no increase in parameters or computations.
    \item A method that easily integrates into existing BN implementations.
    \item An online BN-based input data normalization process.
    \item A statistical schema for reporting/evaluating comparative results that eliminates subjectivity and improvements that could be attributed to randomness.
\end{enumerate}

\section{Related Work} \label{related_work}
BN contains operations for both the normalization and transformation of features. Multiple works have proposed related techniques for inter-network feature normalization and/or transformation to achieve improved performance.

Various types of normalization layers exist and differ on how they select which features to group for normalization. Layer normalization (LN) \cite{ba2016layer} applies to each instance across all input feature channels and is commonly employed in transformer networks \cite{vaswani2017attention,dosovitskiy2020image}. Instance normalization (IN) \cite{ulyanov2016instance} is computed in a similar manner, except it normalizes each channel individually for each instance (example) and is frequently used in image-to-image translation \cite{zhu2017unpaired}. Group normalization (GN) \cite{wu2018group} falls between LN and IN by normalizing groups of channels in individual instances. When the group size equals the total number of channels, GN becomes LN. Similarly, when each channel is a separate group, GN becomes IN. The aforementioned normalization layers were designed to perform better than BN in situations where the training batch size is limited. However, BN remains dominant in the context of CNNs and image classification due to the ability of having large batch sizes on common datasets \cite{he2016deep,xie2017aggregated,sandler2018mobilenetv2,tan2019efficientnet}. In our work, we focus on improving the affine transformation component in BN, though our work could be applied to any of the normalization layer types. 

Related work has appeared to extend the affine transformation component of BN. Rather than using only learnable scale and shift parameters, the approach of \cite{jia2019instance} predicts scale and shift values using a small autoencoder network and combines them with the standard learnable versions. Similarly, in \cite{li2020attentive} they combine predicted and learned scale and shift parameters by utilizing a mixture of affine transformations and a feature attention mechanism to obtain the final transformation. A convolutional layer is used to replace the affine transformation in \cite{xu2020batch}. In \cite{liang2020instance}, an attention-based transformation is introduced to integrate instance-specific information into an extra scale parameter. In \cite{gao2021representative}, two different calibration mechanisms are proposed and integrated into BN to calibrate features and to incorporate instance-specific statistics. These mechanisms consist of an initial centering operation that occurs before normalization and a scaling operation in the affine transformation. Their scaling is similar to \cite{liang2020instance}, differing by using the \textit{normalized} features rather than the \textit{raw} input features for the instance-specific information, as well as employing a different initialization. In \cite{arpit2016normalization}, it is mentioned that the initial value of the scale parameter could be treated as a hyperparameter, though not experimented or discussed further. Alternative initialization values for the shift parameter and numerical stability constant ($\epsilon$) are explored in \cite{yang2018a}.

The approaches of \cite{jia2019instance,li2020attentive,liang2020instance,gao2021representative} all initialize the existing BN affine transformation parameters to the standard values of $\gamma$ = 1 and $\beta$ = 0, but employ different initializations for their additional parameters. In \cite{jia2019instance,li2020attentive}, their additional parameters are initialized using samples from a normal distribution, thus their initial scale values are not constant across the network and could have a magnitude $>$1. In \cite{gao2021representative}, they initialize the additional parameters for their scaling operation such that they ``play no role at the beginning of training" \cite{gao2021representative}, however, the resulting initial scale value is actually $<$1. A grid search is performed in \cite{liang2020instance} to select the best performing initialization values, which also produces a resulting initial scale value $<$1, with the ``theoretical understanding of the best initialization [left as] future work" \cite{liang2020instance}. For both \cite{liang2020instance,gao2021representative}, the resulting scale value is $<$1, constant across all BNs in the network, and fixed for any scenario. The approaches of \cite{liang2020instance,gao2021representative} are most similar to our work, but unlike these methods, our approach does not introduce any additional parameters or computations and instead focuses directly on the initialization and updating of the existing scale parameter $\gamma$ for each BN. 

\section{Framework}
In this section, we initially review the BN formulation then describe the two main tenets of our approach: BN scale initialization and learning rate reduction. We then derive the influence of BN on the backward gradients and discuss various aspects that are influenced by the inclusion of BN. Lastly, we introduce a new BN for input data normalization.

\subsection{BN Formulation} \label{formulation}
BN is composed of two sequential channel-wise operations. The first operation, the ``head", normalizes the data. The second operation, the ``tail", performs an affine transformation on the normalized data. Both operations are applied to each individual channel/neuron across a batch of data. 

In the forward pass during training, the head operation first obtains the mean ($\mu_B$) and variance ($\sigma_B^2$) of an incoming batch $B$ of data/features containing $m$ examples ($X = \{x_i\}^m_1$) using  $\mu_B = \frac{1}{m} \sum\nolimits_{i=1}^{m} x_i$ and $\sigma_B^2 = \frac{1}{m} \sum\nolimits_{i=1}^{m} (x_i - \mu_B)^2$. With these statistics, the input batch is then normalized to have zero mean and unit standard deviation using
\begin{equation} \label{eqn:normalization}
    \hat{X} = \frac{X - \mu_B}{\sqrt{\sigma_B^2 + \epsilon}}
\end{equation}
\noindent where $\epsilon$ is a small value used for numerical stability. Throughout training, $\mu_B$ and $\sigma_B^2$ are employed to update momentum-based averages of the overall mean and variance, which serve as global statistics at test time. Last, the tail affine transformation $Y = \gamma\cdot\hat{X} + \beta$ is applied to the normalized data ($\hat{X}$) using learnable scale ($\gamma$) and shift ($\beta$) parameters (again, one for each channel).

\begin{figure}[t]
    \centering
    \includegraphics[width=8cm]{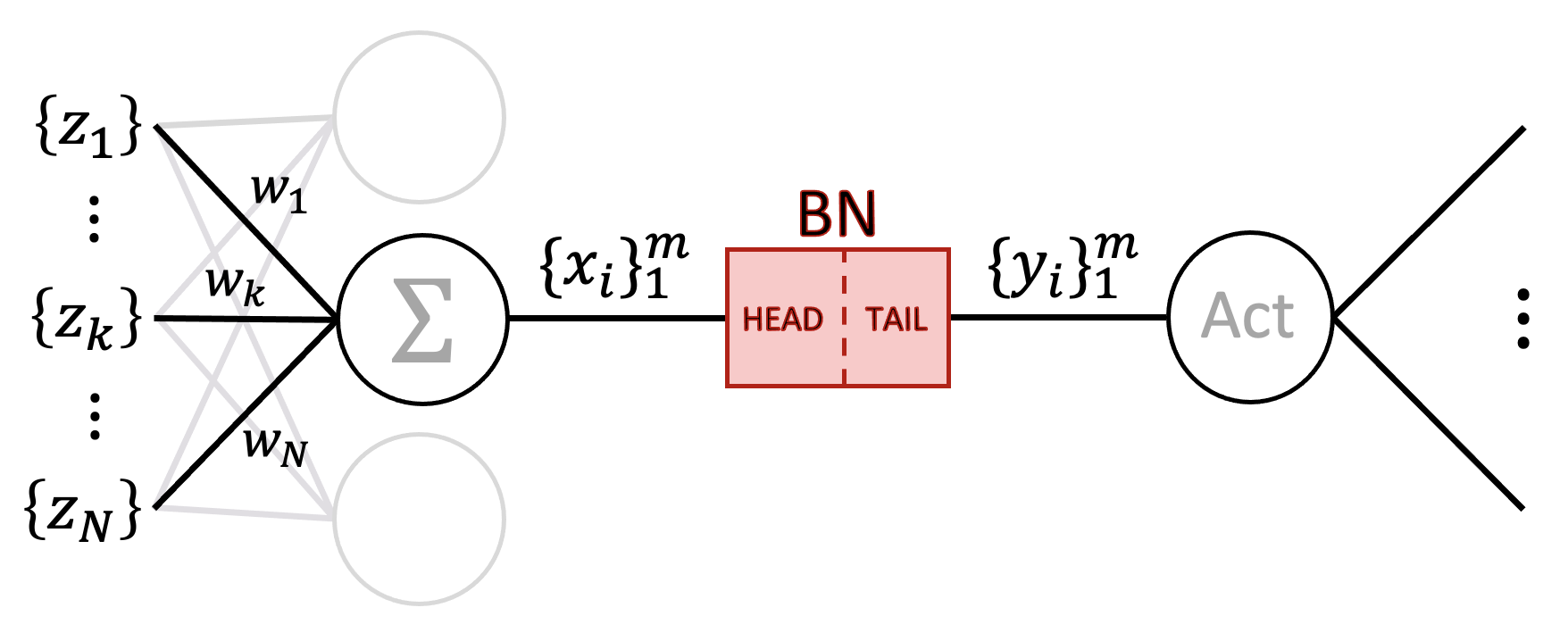}
    \caption{Flow diagram for a single channel/neuron of BN}
    \label{fig:fc_bn_act}
\end{figure}

The data flow when using BN is shown in Fig.~\ref{fig:fc_bn_act}, where BN is typically placed between a linear operator and an activation function. The figure depicts a fully-connected layer for ease in explanation and is easily extended to CNNs.

\subsection{Proposed Adjustments to $\gamma$} \label{adjustments}

The standard procedure for initializing each BN is setting $\gamma$ = 1 and $\beta$ = 0, with the expectation that the affine transformation can learn optimal values and could undo the preceding normalization if desired \cite{ioffe2015batch}. Instead, as we have observed, the learned BN parameters tend to remain close to their initial values which are not likely to be optimal. Furthermore, we have noticed that the BN normalization head (Eqn.~\ref{eqn:normalization}) can often produce overly large values (\eg\ $>$6$\sigma$) as input to the BN affine tail and subsequent convolutional/fully-connected layer (all before another BN), which can be undesirable for learning \cite{lecun-98,lecun-98b,ioffe2015batch}. Therefore, there could exist more optimal values of $\gamma$ not achieved with current learning strategies.

A possible direct solution based on these observations would be to initialize $\gamma <$ 1 such that the data is immediately scaled down at the start of training, addressing the presence of overly large values, and furthermore enabling the shift parameter $\beta$ to have a broader reach on the scaled data before the activation function. Therefore, we propose to treat $\gamma$ as a hyperparameter (for all BNs in the network) to be initialized in the interval $(0,1]$ while leaving the default initialization of $\beta$ = 0.  We will show later that there exists a range of $\gamma$ initialization values within the interval that can lead to significant increases in test accuracy over the default $\gamma = 1$ initialization. 

Since we observed that the learned BN parameters tend to remain somewhat close to their initialization (regardless of the initial value and other hyperparameter settings), it seems necessary to also consider altering the update strategy of these parameters. A seemingly intuitive solution would be to increase the learning rate ($\alpha$) for the BN parameters only, thus allowing larger updates and the ability to potentially explore more optimal values in the parameter space. However, we have seen this {\it degrade} performance. Instead, we propose to \textit{reduce} the learning rate for only the scale $\gamma$ to enable a more fine-grained search, leaving the shift $\beta$ with larger updates to have a broader and more stable search of the normalized and $\gamma$-scaled data. More specifically, we apply a learning rate reduction on $\gamma$ using $\alpha_\gamma = (\alpha / c)$, where $c$ is a positive constant. Though the value of $c$ could be considered another hyperparameter, we experimented with a wide range of values for $c$ and found that performance gains can be achieved as long as the learning rate for $\gamma$ is sufficiently and reasonably reduced. Thus, we {\em fix} this $\gamma$ learning rate reduction {\em constant} to $c$ = 100 for {\em all} experiments.

With a proposed smaller initial value for $\gamma$, it is important to theoretically understand any effect it will have on the gradients for learning. We next examine the gradients of BN to show the influence of $\gamma$ and other important relationships.

\subsection{BN Gradients}
To update a network via gradient descent and backpropagation, loss gradients must be able to propagate through the BN module. For example, parameter $w_1$ in Fig.~\ref{fig:fc_bn_act} is updated using gradient descent by
\begin{equation} \label{eqn:gradient_descent_basic}
    w_1^{new} = w_1^{old} - \alpha \cdot \frac{\partial\mathcal{L}}{\partial w_1}
\end{equation}
\noindent Using the chain rule with a batch of data $\{x_i\}_1^m$, the loss with respect to $w_1$ is computed as
\begin{equation} \label{eqn:w1_loss}
    \frac{\partial\mathcal{L}}{\partial w_1} = \frac{1}{m}\sum_{i=1}^m\frac{\partial\mathcal{L}}{\partial x_i} \cdot \frac{\partial x_i}{\partial w_1}
\end{equation}
\noindent The BN input $x_i$ is computed from the linear operation $x_i = \sum\nolimits_{k=1}^N w_k \cdot z_{k, i}$, where $z_{k, i}$ is the $i^{\text{th}}$ batch value from the $k^{\text{th}}$ channel output from the previous layer (see Fig.~\ref{fig:fc_bn_act}). Therefore $\partial x_i / \partial w_1 = z_{1, i}$ and Eqn.~\ref{eqn:w1_loss} becomes
\begin{equation}
    \frac{\partial\mathcal{L}}{\partial w_1} = \frac{1}{m}\sum_{i=1}^m\frac{\partial\mathcal{L}}{\partial x_i} \cdot z_{1,i}
\end{equation}

Similarly, the gradient $\partial\mathcal{L}/\partial x_i$ through the BN is computed using the chain rule (described in \cite{ioffe2015batch}) as
\begin{equation} \label{eqn:gradient_equation}
    \frac{\partial \mathcal{L}}{\partial x_i} = \frac{\partial \mathcal{L}}{\partial \hat{x}_i} \frac{\partial \hat{x}_i}{\partial x_i} 
    +  \frac{\partial \mathcal{L}}{\partial \sigma_B^2} \frac{\partial \sigma_B^2}{\partial x_i} 
    +  \frac{\partial \mathcal{L}}{\partial \mu_B} \frac{\partial \mu_B}{\partial x_i} 
\end{equation}
\noindent with the $x_i$ dependent gradients
\begin{align}
    \frac{\partial \hat{x}_i}{\partial x_i} = \frac{1}{\sqrt{\sigma_B^2 + \epsilon}} \quad\quad\quad
    \frac{\partial \sigma_B^2}{\partial x_i} = \frac{2}{m} \left( x_i - \mu_B \right) \quad\quad\quad
    \frac{\partial \mu_B}{\partial x_i} = \frac{1}{m} \label{eqn:x_i_derivates} 
\end{align}
\noindent and the remaining gradients
\begin{align}
    \frac{\partial \mathcal{L}}{\partial \hat{x}_i} & =  \frac{\partial \mathcal{L}}{\partial y_i} \cdot \gamma  \label{eqn:bn_loss_wrt_xhat} \\
    \frac{\partial \mathcal{L}}{\partial \sigma_B^2} & =   \sum_{i=1}^{m} \frac{\partial \mathcal{L}}{\partial \hat{x}_i} \left( x_i - \mu_B \right)\cdot\frac{\text{-}1}{2}\left(\sigma_B^2 + \epsilon\right)^{\text{-}\frac{3}{2}} \label{eqn:bn_loss_wrt_var} \\
    \frac{\partial \mathcal{L}}{\partial \mu_B} & =  ( \sum_{i=1}^{m} \frac{\partial \mathcal{L}}{\partial \hat{x}_i} \cdot \frac{1}{\sqrt{\sigma_B^2 + \epsilon}} )+ \frac{\partial \mathcal{L}}{\partial \sigma_B^2} \cdot \frac{2}{m}\sum_{i=1}^{m}\left(x_i - \mu_B \right) \label{eqn:bn_loss_wrt_mean} 
\end{align}
\noindent where $\partial\mathcal{L} / \partial y_i$ is the gradient accumulated from all downstream layers. For thoroughness, the gradients for the affine transformation parameters $\gamma$ and $\beta$ are
\begin{align}
    \frac{\partial\mathcal{L}}{\partial\gamma} = \sum_{i=1}^m \frac{\partial\mathcal{L}}{\partial y_i} \cdot \hat{x}_i \quad \quad \quad \quad \frac{\partial\mathcal{L}}{\partial\beta} = \sum_{i=1}^m \frac{\partial\mathcal{L}}{\partial y_i}
\end{align}

Substituting Eqns.~\ref{eqn:x_i_derivates}-\ref{eqn:bn_loss_wrt_mean} into Eqn.~\ref{eqn:gradient_equation}, we show that 
\begin{align} \label{eqn:substituted_gradient_equation}
    & \frac{\partial \mathcal{L}}{\partial x_i} =  \left( \frac{\partial \mathcal{L}}{\partial y_i} \cdot \gamma \right) \left( \frac{1}{\sqrt{\sigma_B^2 + \epsilon}} \right) + \nonumber \\
    & \left( \sum_{i=1}^{m} \frac{\partial \mathcal{L}}{\partial \hat{x}_i} \left( x_i - \mu_B \right)\cdot\frac{\text{-}1}{2}\left(\sigma_B^2 + \epsilon\right)^{\text{-}\frac{3}{2}} \right)\left( \frac{2}{m} \left( x_i - \mu_B \right) \right) + \nonumber \\
    & \left( ( \sum_{i=1}^{m} \frac{\partial \mathcal{L}}{\partial \hat{x}_i} \cdot \frac{1}{\sqrt{\sigma_B^2 + \epsilon}} ) + \frac{\partial \mathcal{L}}{\partial \sigma_B^2} \cdot  \frac{2}{m}\sum_{i=1}^{m}\left(x_i - \mu_B \right) \right) \left( \frac{1}{m} \right)
\end{align}
\noindent and after multiple reduction steps (novel to our work), 
\begin{align}
    \frac{\partial \mathcal{L}}{\partial x_i} = &\ \frac{\gamma}{\sqrt{\sigma_B^2 + \epsilon}} \cdot \left[   \frac{\partial \mathcal{L}}{\partial y_i} - \frac{1}{m}  \sum_{i=1}^{m} \frac{\partial \mathcal{L}}{\partial y_i}  \left(1 -  \hat{x}_i\right)\right] = \frac{\gamma}{\sqrt{\sigma_B^2 + \epsilon}} \cdot G_i \label{eqn:final_gradient}
\end{align}
\noindent where $G_i$ is a function of the downstream gradients of $y_i$ (\textit{after} the BN layer) and the normalized values $\hat{X}$ (\textit{within} the BN layer). We refer to the leading ratio $\gamma / \sqrt{\sigma_B^2 + \epsilon}$ as the BN `gradient factor', which is composed of the affine scale parameter $\gamma$ for this current BN and the variance $\sigma_B^2$ of the \textit{incoming} data to that BN. Therefore the update for $w_1$ (Eqn.~\ref{eqn:gradient_descent_basic}) is computed as
\begin{equation}
    w_1^{new} = w_1^{old} - \alpha\cdot\frac{\gamma}{\sqrt{\sigma_B^2 + \epsilon}}\left( \frac{1}{m}\sum_{i=1}^m G_i \cdot z_i \right)
\end{equation}
\noindent We will next show how this gradient factor relates to various aspects of learning.

\smallskip
\noindent \textbf{Gradient Influence of $\gamma$.}
Following Fig.~\ref{fig:fc_bn_act} at the beginning of training \textit{before any updates}, the incoming data variance $\sigma_B^2$ to the BN is
\begin{equation} \label{eqn:show_sigmaB}
    \sigma_B^2 = \text{Var}(\{x_i\}^m_1) = \sum_{k=1}^N w_k^2 \cdot \text{Var}(\{z_{k, i}\}_1^m) = \sum_{k=1}^N w_k^2 \cdot \sigma_k^2
\end{equation}
\noindent As each of the $k$ input batches $\{z_{k,i}\}^m_1$ come from the output of a \textit{previous} BN (each normalized then scaled with $\gamma_{prev}$) that is also passed through an activation function,
\begin{equation} \label{eqn:expand_sigma}
    \sum_{k=1}^N w_k^2 \cdot \sigma_k^2 = \sigma^2 \sum_{k=1}^N w_k^2 = \sigma_{act}^2 \cdot \gamma_{prev}^2 \sum_{k=1}^N w_k^2 = \sigma_{act}^2 \cdot \gamma_{prev}^2 \cdot \omega
\end{equation}
\noindent where $\sigma_{act}^2$ is the variance of the normalized (but unscaled) data after being passed through an activation function. For example, a unit Gaussian passed through a ReLU yields a \textit{rectified} normal distribution with an empirical $\sigma_{act}^2 \approx$ (0.58)$^2$. When the normalized data is scaled by $\gamma$ and passed through an activation, the resulting variance is $\sigma^2$ =  $\sigma_{act}^2 \cdot \gamma^2$. 

From Eqn.~\ref{eqn:expand_sigma}, the scale parameter $\gamma_{prev}$ of the \textit{previous} layer BNs is therefore embedded in the \textit{incoming} data variance ($\sigma_B^2$). At the start of training, where $\gamma_{prev}$ = $\gamma_{curr}$, the scales will essentially cancel in the gradient factor of Eqn.~\ref{eqn:final_gradient}
\begin{equation} \label{eqn:constant}
    \frac{\gamma_{curr}}{\sqrt{\sigma_B^2 + \epsilon}} = \frac{\gamma_{curr}}{\gamma_{prev}\cdot\sqrt{\sigma_{act}^2\cdot\omega + \epsilon}} = \frac{1}{\sqrt{\sigma_{act}^2\cdot\omega + \epsilon}}
\end{equation}

\noindent again assuming a negligible $\epsilon$. Thus for the initial backward pass, the initial BN scale value for $\gamma$ has \textit{no effect} on the local BN gradients (the first BN layer will be addressed in Sect.~\ref{input_normalization}). While $\gamma$ will not affect the initial backward pass, it will contribute to the forward pass and loss. Though the value of each $\gamma$ will naturally migrate toward a more optimal value during training, each BN gradient factor should remain non-degenerate. Therefore initializing $\gamma <$ 1 will not cause any cascading issues during training.

\smallskip
\noindent \textbf{Network Weight Initialization.}
Most network weight initialization schemes take into account the number of either incoming or outgoing links in a layer. For example, Kaiming normal initialization \cite{he2015delving} initializes the weights in a network following a normal distribution $\mathcal{N}$(0, $\sigma^2$) with $\sigma = {\text{gain}}/{\sqrt{\text{fan\_mode}}}$, where `gain' is a constant determined by the activation function used in the network and `fan\_mode' is a value representing either the number of incoming (fan-in) or outgoing (fan-out) links. Thus the number of links affects the standard deviation of the normal distribution, with more links yielding a tighter distribution. 

As shown in Eqns.~\ref{eqn:show_sigmaB} and \ref{eqn:expand_sigma}, the variance $\sigma_B^2$ of the data going into a BN at the start of training is dependent on the sum of the squared \textit{incoming} weights ($\omega$). Therefore, weight initialization approaches based on \textit{fan-in} will initially balance the gradient factor for each BN across the network. In our experiments we therefore employed fan-in weight initialization in our networks.

\subsection{BN-Based Input Normalization} \label{input_normalization}

Traditional methods for normalizing the input data typically utilize either ``fixed" bounds based on the data type min/max or compute ``offline" dataset statistics:

\begin{itemize}[noitemsep,nolistsep]
    \item Fixed: Chooses the mean ($\mu$) as the middle value of the data type and the standard deviation ($\sigma$) as the +/- maximum possible value of the mean-subtracted data (\eg\ $\mu$ = $\sigma$ = 128 for an 8-bit [0-255] image). 
    \item Offline: Preprocess the data to estimate the $\mu$ and $\sigma$ based on the average of averages across batches or individual examples in the dataset.
\end{itemize}

\noindent These methods normalize the data using ($X - \mu$) / $\sigma$ (as used in Eqn.~\ref{eqn:normalization} for BN). 

The fixed input normalization technique can be problematic for images not spanning the full possible range of values for its data type. For example, with a dataset of low-contrast imagery the true $\mu$ and $\sigma$ would be much smaller. The offline statistics approach obviously requires the extra steps of computing the dataset $\mu$ and $\sigma$ before training.

Consider the BN gradient factor $\gamma / \sqrt{\sigma_B^2 + \epsilon}$ for the weights just before the \textit{first} BN in the network on the \textit{initial} backward pass. Here the variance $\sigma_B^2$ is computed from the input data itself (Eqn.~\ref{eqn:show_sigmaB}). Examining Eqn.~\ref{eqn:expand_sigma}, we expect this variance to be $\sigma_{act}^2 \cdot \gamma^2 \cdot \omega$ as for all other layers, however neither the scale $\gamma$ or activation are present. Therefore, a \textit{smaller}/\textit{larger} $\sigma^2$ of the input data, where $\sigma^2 \neq \sigma_{act}^2\cdot\gamma^2\cdot\omega$, will result in a \textit{larger}/\textit{smaller} gradient factor as compared with the other layers in the network. 

This bias with traditional input normalization techniques can be removed by conveniently prepending a new BN layer to the network. Since it would not be ideal to immediately threshold the input data with an activation following this new BN layer, we initialize these BNs to have a modified scale value $\hat{\gamma}$ = $\sigma_{act} \cdot \gamma$, where $\hat{\gamma}$ = $0.58\cdot\gamma$ for a ReLU-based network, to account for the missing activation function's influence on the variance. For this BN layer, the built-in shift parameter $\beta$ is unnecessary as no activation function follows and can therefore be removed or fixed to $\beta$ = 0. Thus, employing this BN-based input data normalization method will only add a single scale parameter $\hat{\gamma}$ per input channel (\eg\ 3 parameters total for an RGB image in a CNN).

\section{Experiments} \label{experiments}
To evaluate our proposed approach, we conducted a series of classification experiments using several datasets and network architectures, examined different possible versions of our approach, and compared to relevant existing methods.

\subsection{Datasets \& Network Architectures} \label{dataset_details}
We employed four established classification datasets for our evaluation. CIFAR-10 \cite{krizhevsky2009learning} contains 10 classes, each with 5K and 1K images for training and testing, respectively. The larger, yet distinct, CIFAR-100 \cite{krizhevsky2009learning} has 100 classes, each with 500 training and 100 testing examples. The fine-grained visual classification (FGVC) dataset CUB-200 \cite{WahCUB_200_2011} contains 200 classes with 5994 training and 5794 testing examples. Finally, Stanford Cars (ST-Cars) \cite{KrauseStarkDengFei-Fei_3DRR2013} is another FGVC dataset that consists of 196 classes with 8144 and 8041 images for training and testing, respectively. In order to have a validation set for each of the datasets, we randomly sampled 10\% of the training examples class-wise for validation. 

We primarily employed networks from the ResNet \cite{he2016deep} architecture family, using ResNet-18 for CIFAR-10, ResNet-34 for CIFAR-100, and ResNet-50 for CUB-200/ST-Cars. Due to the smaller image sizes of CIFAR-10/CIFAR-100, common modifications to its associated model were made which consisted of altering the first convolutional layer to have a 3x3 kernel with stride of 1 and padding of 1 (originally a 7x7 with stride of 2 and padding of 2) and removing the max pooling operation that followed the initial convolutional layer. For initialization, convolutional layer weights used Kaiming normal initialization \cite{he2015delving} with `fan-in' mode for a ReLU nonlinearity, fully connected layer weights were initialized with a uniform distribution (also using `fan-in'), and all biases (including the BN shift parameter $\beta$) were initialized to 0. The BN scale parameter $\gamma$ will be initialized with various values to demonstrate our approach. We additionally examined RepVGG \cite{ding2021repvgg}, MobileNetV2 \cite{sandler2018mobilenetv2}, and ResNeXt \cite{xie2017aggregated} architectures (with the same initialization strategy).

\subsection{Training Details} \label{experimental_details}
Training is implemented using a standard regime and in a manner where all hyperparameters are set using typical values. Across all experiments, we used SGD with momentum (0.9) and weight decay (1e-4) on the convolutional and fully connected layer weights. BN parameters and all network biases are properly excluded from weight decay as it is not necessary to impose a constraint to minimize these parameter values. Networks were each trained for 180 epochs with a half-period cosine learning rate scheduler and across multiple initial learning rates (to be presented). We used a batch size of 128 for CIFAR-10/CIFAR-100 and 64 for CUB-200/ST-Cars. Data augmentation schemes for CIFAR-10/CIFAR-100 consisted of random horizontal flipping only and for CUB-200/ST-Cars images were resized to 256x256 then randomly cropped to 224x224, followed by random horizontal flipping. After augmentations, the input data is normalized online using our proposed BN-based input normalization technique with the initialization of $\hat{\gamma}$ = $0.58\cdot\gamma$ corresponding to ReLU activations (Sect.~\ref{input_normalization}). In our experiments, the epoch yielding the best validation accuracy was used to select the final model. Our approach was implemented using PyTorch \cite{paszke2019pytorch} and all models were trained and evaluated on a single NVIDIA V100 GPU.

We note that we {\em purposefully} used minimal regularization techniques in our experiments to show noticeable improvements that can be attributed solely to changes within BN using our approach. Our goal is {\em not} to provide new SOTA scores on the datasets used for evaluation. However, if more regularization is employed to improve baseline performance, our approach may still provide further gains (as shown in an experiment below).

To demonstrate our proposed BN scale initialization approach, we explored a variety of possible $\gamma$ values in the half-open unit interval (0, 1] and several learning rates $\alpha$. Smaller values of $\gamma$ and larger values of $\alpha$ are typically favored and thus we examined a subset of values for $\gamma$ and $\alpha$ after the initial CIFAR-10 experiments. As described in Sect.~\ref{adjustments}, we also divided $\alpha$ by $c$ = 100 for only $\gamma$. For our baseline comparison, referred to as `BASE', we used the default BN scale initialization ($\gamma$ = 1) and no learning rate reduction ($c$ = 1), but retained the BN-based input normalization for fair comparison.

\subsection{Statistical Significance}
As variations in the final score (accuracy) can be caused by different random number generator (RNG) seeds resulting in different network weight initializations and different batching of training data \cite{picard2021torch,wightman2021resnet}, we conducted several runs for each experiment and performed a statistically-grounded comparative analysis on our results. For \textit{each} experiment (a specific BN scale value $\gamma$ and learning rate $\alpha$), we trained 15 networks (of the same form), each with a different RNG seed, and reported the mean and standard deviation of test accuracy. Our seeds were sequential numbers in the range [1, 15], which were made more complex using MD5 as suggested in \cite{jones_pseudo,PyTorchGenerator}. 

A result is judged to be significantly better than BASE according to a one-sided paired t-test \cite{devore} with a significance level of $p \leq 0.05$ using the scores from the 15 runs. All significant improvements over BASE are emphasized in {\bf bold} in the following tables and the highest mean for each experiment is \underline{underlined}. We suggest this method as a proper technique to evaluate and compare empirical results to instill confidence for the reader on reported improvements.

\subsection{Results} \label{base}
We first conducted experiments on CIFAR-10 using the selected BN scale initialization values, then compared our method with possible alternative BN scale formulations. Next, we evaluated our method on different datasets and network architectures, and then compared with different methods of input normalization. Finally, we compared to relevant established approaches on all of the datasets.

\begin{table}[t]
\scriptsize
\setlength\extrarowheight{0.5pt}
\begin{center}
\caption{Accuracy on (a) CIFAR-10 and (b) CIFAR-100, CUB-200, and ST-Cars}
\begin{subtable}[t]{0.45\textwidth}
\caption{}
\begin{tabular}{c|| P{1.5cm}P{1.5cm}P{1.5cm} }
\hline
& \multicolumn{3}{c}{{Learning Rate ($\alpha$)}} \\
$\gamma$ & 0.1 & 0.01 & 0.001 \\ \hhline{=||===}
0.01 & 85.50\textcolor{black!65}{\tiny$\pm$0.39} & {\bf 87.11}\textcolor{black!65}{\tiny$\pm$0.23}  & \underline{\bf 80.37}\textcolor{black!65}{\tiny$\pm$0.58}  \\
0.05  & {\bf 90.19}\textcolor{black!65}{\tiny$\pm$0.32}   & \underline{\bf88.84}\textcolor{black!65}{\tiny$\pm$0.32}  & {\bf 76.98}\textcolor{black!65}{\tiny$\pm$0.71}  \\
0.10  & \underline{\bf 90.80}\textcolor{black!65}{\tiny$\pm$0.20}   & {\bf87.31}\textcolor{black!65}{\tiny$\pm$0.37}  & {\bf 74.48}\textcolor{black!65}{\tiny$\pm$0.55}  \\
0.25  & {\bf 90.32}\textcolor{black!65}{\tiny$\pm$0.24}   & {\bf85.33}\textcolor{black!65}{\tiny$\pm$0.43}  & {\bf 73.83}\textcolor{black!65}{\tiny$\pm$0.64}  \\
0.50  & {\bf 90.17}\textcolor{black!65}{\tiny$\pm$0.19}  & 84.60\textcolor{black!65}{\tiny$\pm$0.35}  & {\bf 72.80}\textcolor{black!65}{\tiny$\pm$0.68}  \\
0.75   & {\bf 90.19}\textcolor{black!65}{\tiny$\pm$0.18}   & 84.43\textcolor{black!65}{\tiny$\pm$0.30}  & {\bf 72.01}\textcolor{black!65}{\tiny$\pm$0.58}  \\ 
1.00 & {\bf 89.81}\textcolor{black!65}{\tiny$\pm$0.46} & 84.48\textcolor{black!65}{\tiny$\pm$0.33}  & 71.15\textcolor{black!65}{\tiny$\pm$0.56}  \\ \hline
BASE & 89.44\textcolor{black!65}{\tiny$\pm$0.45} & 84.64\textcolor{black!65}{\tiny$\pm$0.25}  & 71.32\textcolor{black!65}{\tiny$\pm$0.60}  \\ \hline
\end{tabular} 
\label{tab:cifar10}
\end{subtable} \quad\quad
\begin{subtable}[t]{0.45\textwidth}
\caption{}
\begin{tabular}{c||c|| P{1.5cm}P{1.5cm} }
\hline
& & \multicolumn{2}{c}{{Learning Rate ($\alpha$)}} \\
Dataset & $\gamma$ & 0.1 & 0.01 \\ \hhline{=||=||==}
CIFAR100       & 0.05  &  {\bf 68.18}\textcolor{black!65}{\tiny$\pm$0.30}           &  \underline{\bf 64.01}\textcolor{black!65}{\tiny$\pm$0.54}  \\
               & 0.10  &  \underline{\bf 68.80}\textcolor{black!65}{\tiny$\pm$0.49} &  {\bf 62.83}\textcolor{black!65}{\tiny$\pm$0.48}  \\
               & 0.50  &  {\bf 67.74}\textcolor{black!65}{\tiny$\pm$0.44}           &  {\bf 59.05}\textcolor{black!65}{\tiny$\pm$0.41}  \\ \cline{2-4}
               &  BASE &  66.01\textcolor{black!65}{\tiny$\pm$0.95}                 &  58.48\textcolor{black!65}{\tiny$\pm$0.53}  \\ \hline
CUB-200      & 0.05   & {\bf 58.32}\textcolor{black!65}{\tiny$\pm$0.54}            & 34.23\textcolor{black!65}{\tiny$\pm$0.86}  \\
            & 0.10   & \underline{\bf 58.52}\textcolor{black!65}{\tiny$\pm$0.69}  & {39.92}\textcolor{black!65}{\tiny$\pm$0.76}  \\
            & 0.50   & {\bf 50.45}\textcolor{black!65}{\tiny$\pm$1.22}          & \underline{\bf 45.31}\textcolor{black!65}{\tiny$\pm$0.59}  \\\cline{2-4}
            &  BASE  & 46.26\textcolor{black!65}{\tiny$\pm$1.59}                  & 41.61\textcolor{black!65}{\tiny$\pm$1.03}  \\ \hline
ST-Cars         & 0.05   &  \underline{\bf 78.29}\textcolor{black!65}{\tiny$\pm$0.44} &  {44.18}\textcolor{black!65}{\tiny$\pm$1.29}  \\
             & 0.10   &  {\bf 78.26}\textcolor{black!65}{\tiny$\pm$0.61} &  {48.60}\textcolor{black!65}{\tiny$\pm$1.02}  \\
             & 0.50   &  {\bf 70.17}\textcolor{black!65}{\tiny$\pm$1.45} &  {51.18}\textcolor{black!65}{\tiny$\pm$2.16}  \\ \cline{2-4}
             &  BASE  &  {64.73}\textcolor{black!65}{\tiny$\pm$2.87} &  \underline{51.86}\textcolor{black!65}{\tiny$\pm$1.80}  \\ \hline
\end{tabular}
\label{tab:datasets}
\end{subtable}
\end{center}
\end{table}

\smallskip
\noindent \textbf{CIFAR-10.}
 For our initial experiments with CIFAR-10, we used $\gamma \in \{1.0, 0.75, \newline 0.5, 0.25, 0.1, 0.05, 0.01\}$ and $\alpha \in \{0.1, 0.01, 0.001\}$. Results on CIFAR-10 are presented in Table \ref{tab:cifar10}. Using the proposed $\gamma <$ 1 initialization with the learning rate reduction ($c$ = 100), significant improvements are found across all learning rates. These results alone demonstrate that the default settings and learning of the BN affine transformation (BASE) are not ideal. For this dataset, there is a noticeable trend that as the learning rate decreases, a smaller initialization value for $\gamma$ is favored, with smaller values for $\gamma$ preferred in general. Performance dropped heavily for BASE at lower learning rates, however our approach with smaller $\gamma$ initializations had much less of a decrease. We also experimented with initially setting the BN scale parameter to $\gamma >$ 1, but experiments produced degraded results, as expected.

With only the learning rate reduction method ($c$ = 100,\ \ $\gamma$ = 1), there remained a significant gain over BASE ($c$ = 1, $\gamma$ = 1) at the highest learning rate ($\alpha$ = 0.1). To confirm the importance of learning rate reduction, we ablated it from our method (\ie\ $c$ = 1, $\gamma <$ 1) and results showed {\em degraded} performance from the scores presented in Table \ref{tab:cifar10}. Thus, combining our proposed BN scale initialization and learning rate reduction {\em together} is necessary to obtain the best performance.

In additional experiments with a {\em stronger} BASE model (employing RandomCrop data augmentation \cite{he2016deep,Hoffer_2020_CVPR,Zhang_2021_CVPR}) having 93.79\% accuracy, our proposed BN scale initialization at $\gamma = \{1.0, 0.75, 0.5, 0.25, 0.1\}$ and BN scale learning rate reduction at $c=100$ still provided statistically significant improvements.

\smallskip
\noindent \textbf{Alternative Scale Formulation.}
Given the standard affine formulation $y = \gamma\cdot\hat{X}+\beta$ of BN (as we employ), two similar scale-based alternatives could be $\text{A1:}\ \ Y=\gamma(\hat{X} + \beta) \label{eqn:alt1}$ and $\text{A2:}\ \ Y=\gamma(\gamma_0\cdot\hat{X} + \beta) \label{eqn:alt2}$. In all three formulations, initialization is 0 $< \gamma \leq$ 1, $\gamma_0$ = 1, and $\beta$ = 0. The two alternatives differ from the standard formulation in that they apply scaling to $\beta$ as well, which we argue could limit the ability to shift the data sufficiently before the ReLU activation. While A1 does not introduce any extra parameters or additional computations (similar to ours), A2 includes the extra $\gamma_0$ parameter (and is similar to \cite{huang2020layer}).

We compared the three formulations on CIFAR-10 using the same $\gamma$ values and learning rates presented in Table \ref{tab:cifar10}. All approaches gave significant improvements over BASE, further indicating that the default initialization of BN is not appropriate. At the highest learning rate, there was only a slight difference between the scores (within $0.15\%$ of each other). However, at the lower learning rates the standard formulation (ours) had a clear and significant advantage (up to $7\%$ gain). These results further support our argument that scaling down only the normalized data with $\gamma$ enables the shift parameter $\beta$ to act on a broader range for the following activation function.

\smallskip
\noindent \textbf{CIFAR-100, CUB-200, and ST-Cars.}
We next evaluated our approach on CIFAR-100, CUB-200, and ST-Cars. As mentioned in Sect.~\ref{dataset_details}, we employed ResNet-34 for CIFAR-100 and used ResNet-50 for CUB-200/ST-Cars. Networks for CUB-200/ST-Cars typically employ pretrained ImageNet \cite{deng2009imagenet} models and finetune, however since pretrained models utilize the default BN initialization (which affects the BN statistics and network weights learned), we therefore train all of our networks from scratch. The results with $\gamma \in \{0.5, 0.1, 0.05\}$ and $\alpha \in \{0.1, 0.01\}$ are reported in Table \ref{tab:datasets}.

It is clear that our approach with smaller $\gamma$ initialization and reduced $\gamma$ updates can still produce significant improvements in test accuracy for these datasets that contain a larger number of classes than CIFAR-10. For the highest learning rate, all examined scale values produced significantly better results, but at the lower learning rate for ST-Cars it was unable to achieve significant gains with any of the initialization values, though one scale value ($\gamma$ = 0.5) was not significantly different from BASE. One could argue that this lower learning rate is not adequate for ST-Cars.

\smallskip
\noindent \textbf{Network Depth and Architectures.}
We further examined how our approach extends to different ResNet depths and other network architectures on CIFAR-10. In particular, we trained and examined the deeper ResNet-50, ResNet-101, and ResNet-152 models, and additionally employed MobileNetV2 \cite{sandler2018mobilenetv2}, RepVGG-A0 \cite{ding2021repvgg}, and ResNeXt-50 \cite{xie2017aggregated}. Again, since CIFAR-10 consists of small input image sizes, modifications were made to the other architectures. For MobileNetV2, the initial convolutional layer and second and third bottleneck blocks were changed to have a stride of 1. Similarly, for RepVGG-A0, stage 0 and stage 1 were adjusted to have a stride of 1. ResNeXt-50 was modified similar to ResNet (Sect.~\ref{dataset_details}).

\begin{table}[t]
\scriptsize
\setlength\extrarowheight{0.3pt}
\begin{center}
\caption{(a) Accuracy on CIFAR-10 employing various network depths and network architectures. (b) Comparison with RBN and IEBN variations on CIFAR-10, CIFAR-100, CUB-200, and ST-Cars}
\begin{subtable}[t]{0.45\textwidth}
\caption{}
\begin{tabular}{c||c|| P{1.4cm}P{1.4cm} }
\hline
& & \multicolumn{2}{c}{{Learning Rate ($\alpha$)}} \\
Network & $\gamma$ & 0.1 & 0.01 \\ \hhline{=||=||==}
ResNet-50& 0.05  & {\bf 91.23}\textcolor{black!65}{\tiny$\pm$0.20}      & \underline{\bf 89.60}\textcolor{black!65}{\tiny$\pm$0.19}  \\
          & 0.10   & \underline{\bf 91.28}\textcolor{black!65}{\tiny$\pm$0.26}     & {\bf 87.67}\textcolor{black!65}{\tiny$\pm$0.20}  \\
          & 0.50   & {\bf 89.49}\textcolor{black!65}{\tiny$\pm$0.27}     & 84.74\textcolor{black!65}{\tiny$\pm$0.37}  \\ \cline{2-4}
          & BASE  & 86.94\textcolor{black!65}{\tiny$\pm$1.23}            & 85.04\textcolor{black!65}{\tiny$\pm$0.32}  \\ \hline
ResNet-101 & 0.05  & \underline{\bf 91.58}\textcolor{black!65}{\tiny$\pm$0.22}     & \underline{\bf 90.02}\textcolor{black!65}{\tiny$\pm$0.22}  \\
           & 0.10   & {\bf 91.26}\textcolor{black!65}{\tiny$\pm$0.18}    & {\bf 88.35}\textcolor{black!65}{\tiny$\pm$0.28}  \\ 
           & 0.50   & {\bf 89.89}\textcolor{black!65}{\tiny$\pm$0.74}    & {\bf 85.23}\textcolor{black!65}{\tiny$\pm$0.50}  \\ \cline{2-4}
           &  BASE & 88.28\textcolor{black!65}{\tiny$\pm$1.39}           & 84.74\textcolor{black!65}{\tiny$\pm$0.56}  \\ \hline
ResNet-152 & 0.05  & \underline{\bf 91.20}\textcolor{black!65}{\tiny$\pm$0.16}     & \underline{\bf 90.00}\textcolor{black!65}{\tiny$\pm$0.17}  \\
           & 0.10   & {\bf 90.89}\textcolor{black!65}{\tiny$\pm$0.41}    & {\bf 88.31}\textcolor{black!65}{\tiny$\pm$0.33}  \\ 
           & 0.50   & {\bf 90.17}\textcolor{black!65}{\tiny$\pm$0.23}    & {\bf 85.23}\textcolor{black!65}{\tiny$\pm$0.62}  \\ \cline{2-4}
           &  BASE & 88.73\textcolor{black!65}{\tiny$\pm$0.62}           & 84.15\textcolor{black!65}{\tiny$\pm$0.79}  \\ \hhline{=||=||==}
RepVGG-A0 & 0.05  &  89.71\textcolor{black!65}{\tiny$\pm$0.30}           & \underline{\bf 88.08}\textcolor{black!65}{\tiny$\pm$0.26}  \\
          & 0.10   &  \underline{\bf 90.63}\textcolor{black!65}{\tiny$\pm$0.26}    & {\bf 86.67}\textcolor{black!65}{\tiny$\pm$0.28}  \\ 
          & 0.50   &  {\bf 89.96}\textcolor{black!65}{\tiny$\pm$0.20}    & {\bf 84.28}\textcolor{black!65}{\tiny$\pm$0.47}  \\ \cline{2-4} 
          &  BASE &  89.59\textcolor{black!65}{\tiny$\pm$0.27}           & 83.92\textcolor{black!65}{\tiny$\pm$0.38}  \\ \hline
MobileNetV2 & 0.05  &  88.32\textcolor{black!65}{\tiny$\pm$0.21}        &  \underline{\bf 87.02}\textcolor{black!65}{\tiny$\pm$0.38}  \\
             & 0.10   &  \underline{\bf 91.14}\textcolor{black!65}{\tiny$\pm$0.17} &  {\bf 85.10}\textcolor{black!65}{\tiny$\pm$0.30}  \\ 
             & 0.50   &  {\bf 91.07}\textcolor{black!65}{\tiny$\pm$0.21} &  {\bf 81.25}\textcolor{black!65}{\tiny$\pm$0.41}  \\ \cline{2-4}
             &  BASE &  90.60\textcolor{black!65}{\tiny$\pm$0.19}        &  80.03\textcolor{black!65}{\tiny$\pm$0.57}  \\ \hline
ResNeXt-50 & 0.05  &  \underline{\bf 92.04}\textcolor{black!65}{\tiny$\pm$0.22}    &  \underline{\bf 89.06}\textcolor{black!65}{\tiny$\pm$0.24}  \\
           & 0.10   &  {\bf 92.02}\textcolor{black!65}{\tiny$\pm$0.19}   &  {\bf 86.10}\textcolor{black!65}{\tiny$\pm$0.24} \\ 
           & 0.50   &  {\bf 91.21}\textcolor{black!65}{\tiny$\pm$0.47}   &  80.31\textcolor{black!65}{\tiny$\pm$0.53} \\ \cline{2-4}
           &  BASE &  88.60\textcolor{black!65}{\tiny$\pm$1.57}          &  82.60\textcolor{black!65}{\tiny$\pm$0.43}  \\ \hline
\end{tabular}
\label{tab:networks}
\end{subtable}
\quad\quad
\begin{subtable}[t]{0.45\textwidth}
\caption{}
\begin{tabular}{c||c|| P{1.4cm}P{1.4cm} }
\hline
& & \multicolumn{2}{c}{{Learning Rate ($\alpha$)}} \\
Dataset & Method & 0.1 & 0.01  \\ \hhline{=||=||==}
CIFAR10  & RBN & {\bf 90.17}\textcolor{black!65}{\tiny$\pm$0.22} & 84.72\textcolor{black!65}{\tiny$\pm$0.29}   \\
         &  \ RBN$^\text{-}$ & {\bf 90.11}\textcolor{black!65}{\tiny$\pm$0.24} & 84.50\textcolor{black!65}{\tiny$\pm$0.36}   \\
         & IEBN & {\bf 90.18}\textcolor{black!65}{\tiny$\pm$0.26} & {\bf 85.34}\textcolor{black!65}{\tiny$\pm$0.39}   \\
         &  \ IEBN$^\text{-}$ & {\bf 90.15}\textcolor{black!65}{\tiny$\pm$0.24} & {\bf 85.29}\textcolor{black!65}{\tiny$\pm$0.35}   \\
         &  Ours & \underline{\bf 90.80}\textcolor{black!65}{\tiny$\pm$0.20} & \underline{\bf 88.84}\textcolor{black!65}{\tiny$\pm$0.32}   \\ \cline{2-4}
         &  BASE & 89.44\textcolor{black!65}{\tiny$\pm$0.45} & 84.64\textcolor{black!65}{\tiny$\pm$0.25} \\ \hline
CIFAR100     & RBN           & {\bf 66.95}\textcolor{black!65}{\tiny$\pm$0.57} & {\bf 58.95}\textcolor{black!65}{\tiny$\pm$0.42}   \\
             & \ RBN$^\text{-}$     & {\bf 66.82}\textcolor{black!65}{\tiny$\pm$0.55} & {\bf 58.90}\textcolor{black!65}{\tiny$\pm$0.61}   \\
             & IEBN & {\bf 66.94}\textcolor{black!65}{\tiny$\pm$0.39} & {\bf 60.61}\textcolor{black!65}{\tiny$\pm$0.40}   \\
             &  \ IEBN$^\text{-}$ & {\bf 66.95}\textcolor{black!65}{\tiny$\pm$0.32} & {\bf 60.89}\textcolor{black!65}{\tiny$\pm$0.41}   \\
             &  Ours & \underline{\bf 68.80}\textcolor{black!65}{\tiny$\pm$0.49} & \underline{\bf 64.01}\textcolor{black!65}{\tiny$\pm$0.54}   \\ \cline{2-4}
             & BASE          & {66.01}\textcolor{black!65}{\tiny$\pm$0.95} & {58.48}\textcolor{black!65}{\tiny$\pm$0.53}  \\ \hline
CUB-200     & RBN       & 48.68\textcolor{black!65}{\tiny$\pm$1.56} & {\bf 44.68}\textcolor{black!65}{\tiny$\pm$0.59}   \\
            & \ RBN$^\text{-}$ & 47.14\textcolor{black!65}{\tiny$\pm$2.72} & {\bf 43.02}\textcolor{black!65}{\tiny$\pm$1.22}   \\
            & IEBN & {\bf 54.12}\textcolor{black!65}{\tiny$\pm$0.60} & {\bf 44.92}\textcolor{black!65}{\tiny$\pm$0.74}   \\
            &  \ IEBN$^\text{-}$ & {\bf 53.81}\textcolor{black!65}{\tiny$\pm$0.76} & {\bf 44.09}\textcolor{black!65}{\tiny$\pm$0.65}   \\
            &  Ours & \underline{\bf 58.52}\textcolor{black!65}{\tiny$\pm$0.69} & \underline{\bf 45.31}\textcolor{black!65}{\tiny$\pm$0.59}   \\ \cline{2-4}
            & BASE      & 46.26\textcolor{black!65}{\tiny$\pm$1.59}  & 41.61\textcolor{black!65}{\tiny$\pm$1.03}  \\ \hline
ST-Cars     & RBN           & {\bf 68.17}\textcolor{black!65}{\tiny$\pm$1.84}                 & {51.87}\textcolor{black!65}{\tiny$\pm$1.34}   \\
         & \ RBN$^\text{-}$     & {\bf 67.84}\textcolor{black!65}{\tiny$\pm$2.96}                 & \underline{52.30}\textcolor{black!65}{\tiny$\pm$1.73}   \\
         & IEBN & {\bf 73.60}\textcolor{black!65}{\tiny$\pm$0.92} & 51.06\textcolor{black!65}{\tiny$\pm$0.87}   \\
         &  \ IEBN$^\text{-}$ & {\bf 74.04}\textcolor{black!65}{\tiny$\pm$1.55} & 51.08\textcolor{black!65}{\tiny$\pm$0.78}   \\
         & Ours      &  \underline{\bf 78.29}\textcolor{black!65}{\tiny$\pm$0.44} &  {51.18}\textcolor{black!65}{\tiny$\pm$2.16}   \\ \cline{2-4}
         &  BASE         &  {64.73}\textcolor{black!65}{\tiny$\pm$2.87}               &  {51.86}\textcolor{black!65}{\tiny$\pm$1.80}  \\ \hline
\end{tabular}
\label{tab:comparisons}
\end{subtable}
\end{center}
\end{table}

Results are shown in Table \ref{tab:networks} for $\gamma \in \{0.5, 0.1, 0.05\}$ and $\alpha \in \{0.1, 0.01\}$. Our performance increase was even greater with the deeper ResNet models (as compared to ResNet-18), which suggests our method may be particularly beneficial for training networks containing more parameters/layers faster than with the default BN initialization. Furthermore, our approach transferred well to the other network architectures, emphasizing the generality of our $\gamma <$ 1 initialization and learning rate reduction.

\smallskip
\noindent \textbf{Input Normalization.}
We next compared our proposed input data normalization BN layer to the standard fixed and offline methods (Sect.~\ref{input_normalization}). All experiments were conducted on CIFAR-10 and employed our proposed $\gamma <$ 1 initialization and learning rate reduction ($c$ = 100). Therefore, any differences reported are solely due to how the input data was normalized. Results showed that all three methods produced similar results. Therefore, our proposed online BN-based input normalization method can serve as a replacement for the traditional techniques, where it \textit{automatically} handles input normalization and avoids the previously mentioned issues that may stem from employing the other techniques (Sect.~\ref{input_normalization}). 

\subsection{Related Approaches}
Lastly, we examined the existing related methods of Representative BN (RBN) \cite{gao2021representative} and Instance Enhancement BN (IEBN) \cite{liang2020instance} that utilize instance-specific statistics to compute an \textit{extra} scaling parameter (as described in Sect.~\ref{related_work}). For this work, we focused on the scaling aspect of RBN/IEBN for a more direct comparison of methods. The data normalization component for both approaches is the same as ours (which produces $\hat{X}$), but the affine transformation of RBN/IEBN is $Y = \gamma\cdot\mathcal{S}\cdot\hat{X} + \beta$ with $\mathcal{S} = sigmoid(w_v \cdot \text{GAP}(\mathcal{X}) + w_b)$, where GAP is the global average pooling operation, $w_v$ and $w_b$ are \textit{additional} learnable weights (one of each per channel), and $\mathcal{S}$ is the result of their additional scaling method. 

For RBN, $\mathcal{X}$ = $\hat{X}$ (normalized features) and for IEBN $\mathcal{X}$ = $X$ (raw input features). For RBN, the additional parameters $w_v$ and $w_b$ are initialized to 0 and 1, respectively, while IEBN initializes these parameters to $w_v$ = 0 and $w_b$ = -1. The standard BN parameters are initialized to $\gamma$ = 1 and $\beta$ = 0 for both approaches. Thus the resulting  BN has an initial effective scale of $\gamma\cdot\mathcal{S} \approx$ 0.731 for RBN and 0.269 for IEBN. Both approaches have smaller initial scale values (\ie\ $<$1) similar to ours, though these initial values are \textit{fixed} for \textit{all} scenarios. Related to our use of a reduced learning rate for $\gamma$, their use of a sigmoid will also force smaller gradient updates on $w_v$ and $w_b$ (that affect the resulting scale) since the derivative of a sigmoid function is $sigmoid(x)\cdot(1 - sigmoid(x))$. 

We considered two versions of both approaches, one with the instance-specific statistics (RBN, IEBN) and one that removes the instance-specific statistics and $w_v$ (RBN$^-$, IEBN$^-$). This modifies their scaling method to become $\mathcal{S}=sigmoid(w_b)$ and results in the same initial scale value as previously mentioned. Comparisons of RBN, RBN$^-$, IEBN, IEBN$^-$, and BASE with our approach for all of the datasets are presented in Table \ref{tab:comparisons}. We report results for our approach using the best performing $\gamma$ initialization selected from $\gamma \in \{0.5, 0.1, 0.05\}$.

When evaluated against each other using a one-sided paired t-test (with a significance level of $p \leq 0.05$), our approach significantly outperformed RBN/RBN$^-$ and IEBN/IEBN$^-$ across all datasets and learning rates except for ST-Cars at the lower 0.01 learning rate, where no approach was significantly different from another. As mentioned, this learning rate is arguably insufficient for ST-Cars. 

Our work has shown that \textit{flexibility} in the initialization of $\gamma$ coupled with a reduced learning rate is advantageous to achieve larger performance gains as compared to a \textit{fixed} initial scale value for all situations. Furthermore, the minor differences in results between RBN/IEBN and RBN$^-$/IEBN$^-$ suggest that the contribution of instance-specific information is not as important as the scaling itself. However, it may be possible to incorporate instance-specific statistics with our approach for further gains.

\section{Conclusion}
We revisited BN to address the observed issues of learned BN parameters remaining close to initialization and passing forward overly large values from the normalization. We derived and empirically demonstrated across multiple datasets and network architectures that initializing the BN scale parameter $\gamma <$ 1 and reducing the learning rate on $\gamma$ can yield statistically improved performance over the default initialization and update strategy, suggesting that current training strategies are preventing BN from achieving the most optimal $\gamma$ value. The proposed alterations do not structurally change BN (\ie\ no additional parameters) and can easily be applied to existing implementations. Additionally, we presented a special prepended BN to automatically handle input data normalization during training. 

\smallskip
\noindent \textbf{Acknowledgements.} This research was supported by the U.S.\ Air Force Research Laboratory under Contract \#GRT00054740 (Release \#AFRL-2021-3711). We also thank the DoD HPCMP for the use of their computational resources.


%
%
\bibliographystyle{splncs04}
\bibliography{refs}

\begin{thebibliography}{10}
\providecommand{\url}[1]{\texttt{#1}}
\providecommand{\urlprefix}{URL }
\providecommand{\doi}[1]{https://doi.org/#1}

\bibitem{tensorflow2015}
Abadi, M., Barham, P., Chen, J., Chen, Z., et~al.: Tensorflow: A system for
  large-scale machine learning. In: 12th USENIX Symposium on Operating Systems
  Design and Implementation (2016)

\bibitem{arpit2016normalization}
Arpit, D., Zhou, Y., Kota, B., Govindaraju, V.: {Normalization Propagation: A
  Parametric Technique for Removing Internal Covariate Shift in Deep Networks}.
  In: {International Conference on Machine Learning} (2016)

\bibitem{ba2016layer}
Ba, J.L., Kiros, J.R., Hinton, G.E.: {Layer Normalization}. arXiv preprint
  arXiv:1607.06450  (2016)

\bibitem{bjorck2018understanding}
Bjorck, N., Gomes, C.P., Selman, B., Weinberger, K.Q.: {Understanding Batch
  Normalization}. In: {Advances in Neural Information Processing Systems}
  (2018)

\bibitem{deng2009imagenet}
Deng, J., Dong, W., Socher, R., Li, L.J., Li, K., Fei-Fei, L.: {ImageNet: A
  Large-Scale Hierarchical Image Database}. In: {IEEE Conference on Computer
  Vision and Pattern Recognition} (2009)

\bibitem{devore}
Devore, J.L.: {Probability \& Statistics for Engineering and the Sciences}.
  Brooks / Cole, Cengage Learning, eighth edn. (2011)

\bibitem{ding2021repvgg}
Ding, X., Zhang, X., Ma, N., Han, J., Ding, G., Sun, J.: {RepVGG: Making
  VGG-style ConvNets Great Again}. In: {IEEE/CVF Conference on Computer Vision
  and Pattern Recognition} (2021)

\bibitem{dosovitskiy2020image}
Dosovitskiy, A., Beyer, L., Kolesnikov, A., Weissenborn, D., Zhai, X.,
  Unterthiner, T., Dehghani, M., Minderer, M., Heigold, G., Gelly, S.,
  Uszkoreit, J., Houlsby, N.: {An Image is Worth 16x16 Words: Transformers for
  Image Recognition at Scale}. In: {International Conference on Learning
  Representations} (2021)

\bibitem{gao2021representative}
Gao, S.H., Han, Q., Li, D., Cheng, M.M., Peng, P.: {Representative Batch
  Normalization with Feature Calibration}. In: {IEEE/CVF} {Conference on
  Computer Vision and Pattern Recognition} (2021)

\bibitem{he2015delving}
He, K., Zhang, X., Ren, S., Sun, J.: {Delving Deep into Rectifiers: Surpassing
  Human-Level Performance on ImageNet Classification}. In: {IEEE International
  Conference on Computer Vision} (2015)

\bibitem{he2016deep}
He, K., Zhang, X., Ren, S., Sun, J.: {Deep Residual Learning for Image
  Recognition}. In: {IEEE Conference on Computer Vision and Pattern
  Recognition} (2016)

\bibitem{Hoffer_2020_CVPR}
Hoffer, E., Ben-Nun, T., Hubara, I., Giladi, N., Hoefler, T., Soudry, D.:
  {Augment Your Batch: Improving Generalization Through Instance Repetition}.
  In: {IEEE/CVF Conference on Computer Vision and Pattern Recognition} (2020)

\bibitem{huang2017densely}
Huang, G., Liu, Z., Van Der~Maaten, L., Weinberger, K.Q.: {Densely Connected
  Convolutional Networks}. In: IEEE {C}onference on {C}omputer {V}ision and
  {P}attern {R}ecognition (2017)

\bibitem{huang2020layer}
Huang, L., Qin, J., Liu, L., Zhu, F., Shao, L.: {Layer-wise Conditioning
  Analysis in Exploring the Learning Dynamics of DNNs}. In: {European
  Conference on Computer Vision} (2020)

\bibitem{ioffe2015batch}
Ioffe, S., Szegedy, C.: {Batch Normalization: Accelerating Deep Network
  Training by Reducing Internal Covariate Shift}. In: International
  {C}onference on {M}achine {L}earning (2015)

\bibitem{jia2019instance}
Jia, S., Chen, D.J., Chen, H.T.: {Instance-Level Meta Normalization}. In:
  {IEEE/CVF Conference on Computer Vision and Pattern Recognition} (2019)

\bibitem{jones_pseudo}
Jones, D.: {Good Practice in (Pseudo) Random Number Generation for
  Bioinformatics Applications}. Tech. rep., {University College London} (2010)

\bibitem{KrauseStarkDengFei-Fei_3DRR2013}
Krause, J., Stark, M., Deng, J., Fei-Fei, L.: {3D Object Representations for
  Fine-Grained Categorization}. In: {IEEE Workshop on 3D Representation and
  Recognition} (2013)

\bibitem{krizhevsky2009learning}
Krizhevsky, A., Hinton, G., et~al.: {Learning Multiple Layers of Features from
  Tiny Images}  (2009)

\bibitem{lecun-98}
LeCun, Y., Bottou, L., Bengio, Y., Haffner, P.: {Gradient-Based Learning
  Applied to Document Recognition}. Proceedings of the IEEE  (1998)

\bibitem{lecun-98b}
LeCun, Y., Bottou, L., Orr, G., Muller, K.: {Efficient BackProp}. In: Neural
  {N}etworks: Tricks of the {T}rade (1998)

\bibitem{li2020attentive}
Li, X., Sun, W., Wu, T.: {Attentive Normalization}. In: {European Conference on
  Computer Vision} (2020)

\bibitem{liang2020instance}
Liang, S., Huang, Z., Liang, M., Yang, H.: {Instance Enhancement Batch
  Normalization: An Adaptive Regulator of Batch Noise}. In: AAAI {C}onference
  on {A}rtificial {I}ntelligence (2020)

\bibitem{luo2018towards}
Luo, P., Wang, X., Shao, W., Peng, Z.: {Towards Understanding Regularization in
  Batch Normalization}. In: {International Conference on Learning
  Representations} (2019)

\bibitem{paszke2019pytorch}
Paszke, A., Gross, S., Massa, F., Lerer, A., Bradbury, J., Chanan, G., Killeen,
  T., Lin, Z., Gimelshein, N., Antiga, L., et~al.: {Py{T}orch: An Imperative
  Style, High-Performance Deep Learning Library}. In: Advances in {N}eural
  {I}nformation {P}rocessing {S}ystems (2019)

\bibitem{picard2021torch}
Picard, D.: {\texttt{torch.manual\_seed(3407)} Is All You Need: On the
  Influence of Random Seeds in Deep Learning Architectures for Computer
  Vision}. arXiv preprint arXiv:2109.08203  (2021)

\bibitem{PyTorchGenerator}
PyTorch: {Torch Generator}.
  \url{https://pytorch.org/docs/stable/generated/torch.Generator.html},
  accessed: August 2021

\bibitem{sandler2018mobilenetv2}
Sandler, M., Howard, A., Zhu, M., Zhmoginov, A., Chen, L.C.: {MobileNetV2:
  Inverted Residuals and Linear Bottlenecks}. In: {IEEE/CVF Conference on
  Computer Vision and Pattern Recognition} (2018)

\bibitem{santurkar2018does}
Santurkar, S., Tsipras, D., Ilyas, A., Madry, A.: {How Does Batch Normalization
  Help Optimization?} In: {I}nternational {C}onference on {N}eural
  {I}nformation {P}rocessing {S}ystems (2018)

\bibitem{simonyan2014very}
Simonyan, K., Zisserman, A.: {Very Deep Convolutional Networks for Large-Scale
  Image Recognition}. In: {International Conference on Learning
  Representations} (2015)

\bibitem{tan2019efficientnet}
Tan, M., Le, Q.: {EfficientNet: Rethinking Model Scaling for Convolutional
  Neural Networks}. In: {International Conference on Machine Learning} (2019)

\bibitem{ulyanov2016instance}
Ulyanov, D., Vedaldi, A., Lempitsky, V.: {Instance Normalization: The Missing
  Ingredient for Fast Stylization}. arXiv preprint arXiv:1607.08022  (2016)

\bibitem{vaswani2017attention}
Vaswani, A., Shazeer, N., Parmar, N., Uszkoreit, J., Jones, L., Gomez, A.N.,
  Kaiser, {\L}., Polosukhin, I.: {Attention Is All You Need}. In: Advances in
  {N}eural {I}nformation {P}rocessing {S}ystems (2017)

\bibitem{WahCUB_200_2011}
Wah, C., Branson, S., Welinder, P., Perona, P., Belongie, S.: {The Caltech-UCSD
  Birds-200-2011 Dataset}. Tech. rep., {California Institute of Technology}
  (2011)

\bibitem{wightman2021resnet}
Wightman, R., Touvron, H., J{\'e}gou, H.: {ResNet Strikes Back: An Improved
  Training Procedure in timm}. arXiv preprint arXiv:2110.00476  (2021)

\bibitem{wu2018group}
Wu, Y., He, K.: {Group Normalization}. In: {E}uropean {C}onference on
  {C}omputer {V}ision (2018)

\bibitem{xie2017aggregated}
Xie, S., Girshick, R., Doll{\'a}r, P., Tu, Z., He, K.: {Aggregated Residual
  Transformations for Deep Neural Networks}. In: {IEEE Conference on Computer
  Vision and Pattern Recognition} (2017)

\bibitem{xu2020batch}
Xu, Y., Xie, L., Xie, C., Mei, J., Qiao, S., Wei, S., Xiong, H., Yuille, A.:
  {Batch Normalization with Enhanced Linear Normalization}. {arXiv preprint
  arXiv:2011.14150}  (2020)

\bibitem{yang2018a}
Yang, G., Pennington, J., Rao, V., Sohl-Dickstein, J., Schoenholz, S.S.: {A
  Mean Field Theory of Batch Normalization}. In: {International Conference on
  Learning Representations} (2019)

\bibitem{Zhang_2021_CVPR}
Zhang, S., Nezhadarya, E., Fashandi, H., Liu, J., Graham, D., Shah, M.:
  {Stochastic Whitening Batch Normalization}. In: {Proceedings of the IEEE/CVF
  Conference on Computer Vision and Pattern Recognition} (2021)

\bibitem{zhu2017unpaired}
Zhu, J.Y., Park, T., Isola, P., Efros, A.A.: {Unpaired Image-to-Image
  Translation using Cycle-Consistent Adversarial Networks}. In: IEEE
  {I}nternational {C}onference on {C}omputer {V}ision (2017)

\end{thebibliography}
\end{document}